\documentclass{article}
\usepackage{ijcai23}

\usepackage{times}
\usepackage{soul}
\usepackage{url}
\usepackage[hidelinks]{hyperref}
\usepackage[utf8]{inputenc}
\usepackage[small]{caption}
\usepackage{graphicx}
\usepackage{amsmath}
\usepackage{amsthm}
\usepackage{booktabs}
\usepackage{algorithm}
\usepackage{algorithmic}
\usepackage[switch]{lineno}

\usepackage{amsfonts}
\usepackage{subfigure}

\usepackage{setspace}

\title{Generalization through Diversity: Improving Unsupervised Environment Design}

\author{
Submission Number: 4071
}

\author{
Wenjun Li
\and
Pradeep Varakantham 
\and
Dexun Li
\affiliations
Singapore Management University
\emails
\{wjli.2020, pradeepv, dexunli.2019\}@smu.edu.sg
}

\begin{document}

\maketitle

\begin{abstract}
Agent decision making using Reinforcement Learning (RL) heavily relies on either a model or simulator of the environment (e.g., moving in an 8x8 maze with three rooms, playing Chess on an 8x8 board). Due to this dependence, small changes in the environment (e.g., positions of obstacles in the maze, size of the board) can severely affect the effectiveness of the policy learned by the agent. To that end, existing work has proposed training RL agents on an adaptive curriculum of environments (generated automatically) to improve performance on out-of-distribution (OOD) test scenarios. Specifically, existing research has employed the potential for the agent to learn in an environment (captured using Generalized Advantage Estimation, GAE) as the key factor to select the next environment(s) to train the agent. However, such a mechanism can select similar environments (with a high potential to learn) thereby making agent training redundant on all but one of those environments.  To that end, we provide a principled approach to adaptively identify diverse environments based on a novel distance measure relevant to environment design. We empirically demonstrate the versatility and effectiveness of our method in comparison to multiple leading approaches for unsupervised environment design on three distinct benchmark problems used in literature. 
\end{abstract}

\section{Introduction}
Deep Reinforcement Learning (DRL) has had many successes in challenging tasks (e.g., Atari \cite{mnih2015human}, AlphaGO \cite{silver2016mastering}, solving Rubik's cube \cite{akkaya2019solving}, chip design \cite{mirhoseini2021graph}) in the last decade. However, DRL agents have been proven to be brittle and often fail to transfer well to environments only slightly different from those encountered during training \cite{zhang2018dissection,cobbe2019quantifying}, e.g., changing size of the maze, background in a game, positions of obstacles. To achieve robust and generalizing policies, agents must be exposed to different types of environments that assist in learning different challenges~\cite{cobbe2020leveraging}. To that end, existing research has proposed Unsupervised Environment Design (UED) that can create a curriculum (or distribution) of training scenarios that are adaptive with respect to agent policy \cite{mehta2020active,portelas2020teacher,wang2019measuring,dennis2020emergent,jiang2021prioritized,jiang2021replay}. Following the terminology in existing works, we will refer to a particular environment configuration as a \textit{level}, e.g., an arrangement of blocks in the maze, shape/height of the obstacles in front of the agent, the geometric configuration of racing tracks, etc. In the rest of this paper, we will use levels and environments interchangeably. 

\label{related_work_UED}
\subsection{Related Work on UED}
Existing works on UED can be categorized along two threads. The first thread has focused on principled generation of levels and was pioneered by the Protagonist Antagonist Induced Regret Environment Design (PAIRED, \cite{dennis2020emergent}) algorithm. PAIRED introduced a multi-agent game between level {\em generator} (teacher), {\em antagonist} (expert student) and {\em protagonist} (normal student), and utilized the performance difference between {\em antagonist} and {\em protagonist}  as a reward signal to guide the {\em generator} to adaptively create challenging training levels. The second thread has abandoned the idea of principled level generation and instead championed: (a) randomly generating levels and (b) replaying previously considered levels to deal with catastrophic forgetting by the student agent. The algorithm is referred to as Prioritized Level Replay (PLR, \cite{jiang2021prioritized}) and the levels are replayed based on learning potential, captured using Generalized Advantage Estimation (GAE, \cite{schulman2015high}). PLR has been empirically shown to be more scalable and also is able to achieve more robust and better ``out-of-distribution" performance than PAIRED. In \cite{jiang2021replay}, authors combined randomized generator and replay mechanism together and proposed PLR$^\perp$. Empirically, PLR$^\perp$ achieves the state-of-the-art in literature (as a method that demands no human expertise), and thus we will adopt PLR$^\perp$ as our primary baseline in this paper. Another recent approach ACCEL~\cite{parker2022evolving} that builds on PLR performs edits (i.e., modifications based on human expertise) on high-regret environments to learn better. Both these threads of research have significantly improved the state-of-the-art (domain randomization~\cite{sadeghi2016cad2rl,tobin2017domain} and adversarial learning~\cite{pinto2017supervision}). However, to generalize better, it is not sufficient to train the agent only on high-regret/GAE levels. We can have multiple high-regret levels, which are very ``similar" to each other and agent does not learn a lot from being trained on similar levels. Thus, levels also have to be sufficiently ``different", so that agent can gain more perspective on different challenges that it can face. 

To that end, in this paper, we introduce a diversity metric in UED, which is defined based on distance between occupancy distributions associated with agent trajectories from different levels. We then provide a principled method, referred to as \textit{Diversity Induced Prioritized Level Replay} (DIPLR) to select levels with the diversity metric to provide better generalization performance than the state-of-the-art. 

\label{related_work_diversity}
\subsection{Related Work on Diversity in RL}
There is existing literature on diversity measurement in the field of transfer RL, e.g., \cite{song2016measuring,agarwal2021contrastive,wang2019measuring}. However, these works conduct an exhaustive comparison of any possible states between two underlying Markov Decision Processes (MDPs), which does not take the current agent policy into account and may suffer from a large computational cost. 
In contrast, we use the collected trajectories induced by the policy to approximate the occupancy distribution of the different levels.
Such a method can avoid massive computation and more robustly indicate the difference between two levels being explored by the same agent policy. There are a few works that have explored the diversity between policies in DRL.
1) \cite{hong2018diversity} adopted distance on pairwise policies to modify the loss function to enforce the agent to attempt policies different from its prior policies. As an improved version of \cite{hong2018diversity}, \cite{parker2020effective} computes the diversity on a population of policies instead of pairwise distance, so that it avoids cycling training behaviors and policy redundancy. 2) \cite{lupu2021trajectory} proposed trajectory diversity measurement based on \textit{Jensen-Shannon Divergence} (JSD) and included it in the loss function to maximize the difference between policies. The JSD computes the divergence based on action distribution in trajectories. In contrast to these works, our method computes both pairwise and population-wise distance for the purpose of measuring the difference between levels. Furthermore, we provide a diversity-guided method to train a robust and well-generalizing agent in the UED framework.

\label{contribution}
\subsection{Contributions}
Overall, our contributions are three-fold. First, we highlight the benefits of diversity in UED tasks and formally present the definition of distance between levels. Second, we employ Wasserstein distance to quantitatively measure the distance and introduce the DIPLR algorithm. Finally, we empirically demonstrate the versatility and effectiveness of DIPLR in comparison to other leading approaches on benchmark problems.
Notably, we also investigate the relationship between {\em diversity} and {\em regret} (i.e., {\em learning potential}) and conduct an ablation study to reveal their individual effectiveness. 
Surprisingly, we find that diversity solely works better than regret in the model, and they complement each other to achieve the best performance when combined.

\section{Background}
\label{background}
In this section, we describe the problem of UED and also the details of the leading approaches for solving UED.  

\subsection{Unsupervised Environment Design, UED}
The goal of UED is to train a student agent that performs well across a large set of different environments. To achieve this goal, there is a teacher agent in UED that provides a curriculum of environment parameter values to train the student agent to generalize well to unseen levels. 

UED problem is formally described using an Underspecified Partially Observable Markov Decision Process (UPOMDP). It is defined using the following tuple:
\begin{equation*}
    \langle S, A, \Theta, I, O, T, R, \gamma \rangle
\end{equation*}
$S$, $A$ and $O$ are the set of states, actions and observations respectively. $R:S \rightarrow \mathbb{R}$ is the reward function and $\gamma$ is the discount factor. The most important element in the tuple is $\Theta$, which is the set of environment parameters. A particular parameter $\theta \in \Theta$ (can be a vector or sequence of values) defines a level and can impact the reward model, transition dynamics and the observation function, i.e., $R:S \times \Theta \rightarrow \mathbb{R}$, $T:S \times A \times \Theta \rightarrow S$ and $I: S \times \Theta \rightarrow O$. UPOMDP is underspecified, because we cannot train with all values of $\theta$ ($\in \Theta$), as $\Theta$ can be infinitely large. 
The goal of the student agent policy $\pi$ in a UPOMDP is to maximize its discounted expected rewards for any given $\theta \in \Theta$: $$\mathop{\max}\limits_{\pi} V^\theta(\pi)= \mathop{\max}\limits_{\pi} \mathbb{E}_{\pi} V^\theta(\tau) = \mathop{\max}\limits_{\pi} \mathbb{E}_{\pi} \Big[\sum_{t=0}^H r_t^\theta \cdot \gamma^t \Big]$$ where $r^{\theta}_t$ is the reward obtained by $\pi$ in a level with environment parameter $\theta$ at time step $t$. Thus, the student agent needs to be trained on a series of $\theta$ values that maximize its generalization ability on all possible levels from $\Theta$. To that end, we employ the teacher agent. The goal of the teacher agent policy $\Lambda$ is to generate a distribution over the next set of environment parameter values to train the student, i.e., $$\Lambda: \Pi \rightarrow \Delta(\Theta)$$ to achieve good generalization performance, where $\Pi$ is the set of possible policies of the teacher. 

\subsection{Approaches for Solving UED}
In all approaches for solving UED, the student always optimizes a policy that maximizes its value on the given $\theta$. The different approaches for solving UED vary on the method adopted by the teacher agent.  We now elaborate on the $\Lambda$ employed by different approaches. 

Two fundamental approaches to UED are Domain Randomization (DR) \cite{jakobi1997evolutionary,tobin2017domain} and Minimax \cite{morimoto2005robust,pinto2017supervision}. The teacher in DR simply randomizes the environment configurations regardless of the student's policy, i.e.,
$$\Lambda^{DR}(\pi) = {\cal U}(\Theta)$$
where ${\cal U}$ is the uniform distribution over all possible $\theta$ values. 
On the other hand, teacher in Minimax adversarially generates challenging environments to minimize the rewards of the student's policy, i.e.,
$$\Lambda^{MM}(\pi) = \arg \min_{\theta} V^{\theta}(\pi)$$

\noindent The next set of approaches related to PAIRED~\cite{dennis2020emergent,jiang2021replay} rely on regret, which is defined approximately as the difference between the maximum and the mean return of students' policy, to generate new levels: 
$$reg^\theta(\pi) \approx \max_{\tau \sim \pi} V^{\theta}(\tau) - \mathbb{E}_{\tau \sim \pi} V^\theta(\tau)$$
Given this regret, teacher selects policies that maximize regret:
$$\Lambda^{MR}(\pi) = \arg \max_\theta reg^\theta(\pi)$$
From the original paper by \cite{dennis2020emergent}, there have been multiple major improvements~\cite{jiang2021prioritized,parker2022evolving}:
\begin{enumerate}
\item New levels are generated randomly instead of using an optimizer. 
\item Level to be selected at a time step is decided based on an efficient approximation of regret, referred to as {\em positive value loss} (a customized form of Generalized Advantage Estimation (GAE)):
\begin{equation} 
gae^\theta(\pi) = \frac{1}{H} \sum_{t=0}^{H} \max \left( \sum_{k=t}^{H} (\gamma \lambda)^{k-t} \delta_k, 0 \right) \label{eq:gae}
\end{equation}
where $\gamma$ and $\lambda$ are the GAE and MDP discount factors respectively, $H$ is the time horizon and $\delta_k$ is the TD-error at time step $k$.
\item Levels are replayed with a certain probability to ensure there is no catastrophic forgetting. 
\item Edit randomly on generated levels through human-defined edits~\cite{parker2022evolving}.
\end{enumerate}
We will use learning potential, regret and GAE, interchangeably to represent \textit{positive value loss} in the rest of the paper.

\subsection{Wasserstein Distance}
We will propose a diversity measure that relies on distance between occupancy distributions, where we utilize Wasserstein Distance. The problem of optimal transport density between two distributions was initially proposed by \cite{monge1781memoire} and generalized by \cite{kantorovich1942kantorovich}. Wasserstein distance is preferred by the machine learning community among several commonly used divergence measures, e.g., Kullback-Leibler (KL) divergence, because it is 1) non-zero and continuously differentiable when the supports of two distributions are disjoint and 2) symmetric, i.e.,  $\mathcal{W}(\mathcal{P}, \mathcal{Q})$ is equal to $\mathcal{W}(\mathcal{Q}, \mathcal{P})$.

Wasserstein distance is the measurement between probability distributions defined on a metric space $M(d, \mathcal{C})$ with a cost metric $d: \mathcal{C} \times \mathcal{C} \mapsto \mathbb{R}_{+}$: 
\begin{eqnarray} 
\mathcal{W}_{p}(\mathcal{P}, \mathcal{Q}) & = & \left( \mathop{inf}\limits_{\psi \in \Pi(\mathcal{P},\mathcal{Q})} \mathbb{E}_{(x, y)\sim \psi} [d(x, y)^p] \right) ^{1/p} \label{eq:wasserstein} 
\end{eqnarray}
where $\Pi(\mathcal{P}, \mathcal{Q})$ is the set of all possible joint probability distributions between $\mathcal{P}$ and $\mathcal{Q}$, and one joint distribution $\psi \in \Pi(\mathcal{P}, \mathcal{Q})$ represents a density transport plan from point $x$ to $y$ so as to make $x$ follows the same probability distribution of $y$.

\section{Approach: DIPLR}
\label{method}

A key drawback of existing approaches for solving UED is the inherent assumption that levels with high regret (or GAE) all have high learning potential. However, if there are two levels that are very similar to each other, and they both have high regret, training the student agent on one of those levels makes training on the other unnecessary and irrelevant. Given our ultimate goal is to let student agent policy transfer well to a variety of few-shot or zero-shot levels, we utilize diversity. Diversity has gained traction in Deep Reinforcement Learning to improve the generalization performance (albeit in a different problem settings than the one in the paper) of models (\cite{parker2020effective,hong2018diversity,lupu2021trajectory}).
More specifically, we provide
\begin{enumerate}
\item A scalable mechanism for quantitatively estimating similarity (or dissimilarity) between two given levels given the current student policy. This will then be indirectly used to generate diverse levels. 
\item A teacher agent for UED that generates diverse levels and trains student agent in those diverse environments, so as to achieve strong generalization to "out-of-distribution" levels. 
\end{enumerate}

\subsection{Estimating Similarity between Levels }
The objective here is to estimate the similarity between levels given the current student agent policy. One potential option is to encode each level using one of a variety of encoding methods (e.g., Variational Autoencoder) or using the parameters associated with the level and then taking the distance between encodings of levels. However, such an approach has multiple major issues: 
\begin{itemize}
    \item Encoding a level does not account for the sequential moves the student agent will make through the level, which is of critical importance as the similarity is with respect to student policy; 
    \item There can be stochasticity in the level that is not captured by the parameters of the level.  For instance, in Bipedal-Walker, a complex yet well-parameterized UPOMDP introduced by \cite{wang2019paired,parker2022evolving}, has multiple free parameters controlling the terrain. Because of the existence of stochasticity in the level generation process, two levels could be very different while we have near-zero distance measurement given their environment parameter vectors. 
    \item Distance between environment parameters requires normalization in each parameter dimension and is domain-specific.
\end{itemize}

\noindent Since we collect several trajectories within current levels when approximating the regret value, we can naturally get the state-action distributions induced by the current policy. Therefore, 
we propose to evaluate \textbf{\textit{similarity on the different levels}} based on the \textbf{\textit{distance between occupancy distributions of the current student policy}}. The hypothesis is that if the levels are similar, then the trajectories traversed by the student agent within the level will have similar  state-action distribution.  Equation (\ref{eq:prob_traj}) provides the expression for state-action distribution given a policy, $\pi$ and level $l_{\theta_1}$ ($\theta_1$ represents the parameters corresponding to the level):
\begin{align}
\rho^{\pi}_{l_{\theta_1}}(s,a) &= (1-\gamma) \sum_{t=0}^{H} \Big[ Pr(s_t=s,a_t=a | s_0 \sim p_0(.), \nonumber\\
&\hspace{0.4in} s_t \sim p(.|s_{t-1},a_{t-1},\theta_1),a_t \sim \pi(.|s_t) \Big] \label{eq:prob_traj}
\end{align}
Similarly, we can derive $\rho^\pi_{l_{\theta_2}}$ for level, $l_{\theta_2}$ with parameter $\theta_2$. 
Typically, KL divergence is employed to measure the distance between state-action distributions.
$$-D_{KL}(\rho^{\pi}_{l_{\theta_1}}|| \rho^{\pi}_{l_{\theta_2}}) = \mathbb{E}_{(s,a)\sim \rho^\pi_{\theta_1}}\Big[\log \frac{\rho^\pi_{l_{\theta_2}}}{\rho^\pi_{l_{\theta_1}}}\Big]$$

However, KL divergence is not applicable in our setting, because of the lack of transition probabilities, the two different levels can result in two occupancy distributions with disjoint supports. More importantly, KL divergence cannot work without explicit estimates of the occupancy distributions. Therefore, we employ the Wasserstein distance described in Equation (\ref{eq:wasserstein}), which can calculate the distance between two distributions from empirical samples. 


We can write the Wasserstein distance between two occupancy distributions as Equation (\ref{eq:div_traj_prob_1}):
\begin{equation}\label{eq:div_traj_prob_1}
\resizebox{\linewidth}{!}{$
\mathcal{W}(\rho^{\pi}_{l_{\theta_1}}, \rho^{\pi}_{l_{\theta_2}}) 
= \left( \mathop{inf}\limits_{\psi \in \Pi(\rho^{\pi}_{l_{\theta_1}}, \rho^{\pi}_{l_{\theta_2}})} \mathbb{E}_{(\phi_1, \phi_2)\sim \psi} [d(\phi_1, \phi_2)^p] \right) ^{1/p}  
$}
\end{equation}
where $\phi \in (S, A)$ is a sample from the occupancy distribution. By Equation (\ref{eq:div_traj_prob_1}), we can collect state-action samples in trajectories to compute the empirical Wasserstein distance between two levels, i.e., ${\cal D}(l_{\theta_1}, l_{\theta_2}) \approx \mathcal{W}(\rho^{\pi}_{l_{\theta_1}}, \rho^{\pi}_{l_{\theta_2}})$, is our empirical estimation of the Wasserstein distance between two levels.

\subsection{Diversity Guided Training Agent}
\begin{figure}[t]
  \centering
  \subfigure{\includegraphics[width=1\linewidth]{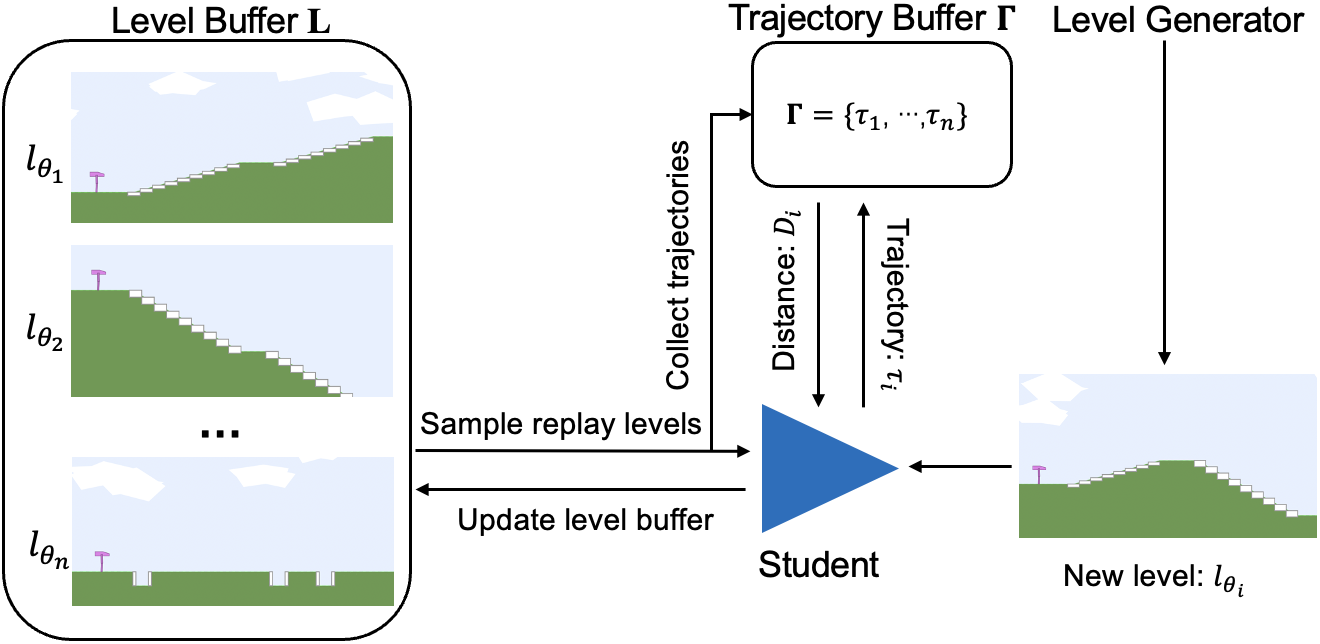}}
  \caption{An Overview of DIPLR Algorithm}
  \label{fig:algo_pipeline}
\end{figure}

Now that we have a scalable way to compute the distance between levels, ${\cal D}(\cdot,\cdot)$, we will utilize it in the teacher to generate a diverse curriculum for the student agent. We build on the PLR training agent~\cite{jiang2021replay} that only considered regret/GAE and staleness to prioritize levels for selection. Our training agent most crucially employs diversity to select levels. 

Our training agent maintains a buffer of high-potential levels for training. At each iteration, we either: (a) generate a new level to be added to the buffer; or (b) sample a mini-batch of levels for the student agent to train on.

\subsubsection{Diversity During Generating a New Level}
For a new level, $l_{\theta}$, the distance from $l_{\theta}$ to all the levels, $L = \left<l_{\theta_1}, l_{\theta_2}, \cdots \right>$ in the buffer is given by:
\begin{align}
{\cal D}(l_{\theta}, L) = \min_{k} {\cal D}(l_{\theta}, l_{\theta_k}) \label{eq:div_level2buffer}
\end{align}
To increase diversity, we add new levels to the buffer that have the highest value of this distance (amongst all the randomly generated levels).  We can also combine this distance measure with regret/GAE when taking a decision to include a level in the buffer so that different and challenging levels are added. 

\subsubsection{Diversity in Sampling Levels from Buffer to Train}
To decide on levels to train on, we maintain a priority (probability of getting selected) for each level that is computed based on their distance from other levels in the buffer. The rank of a level, $l_{\theta_i}$ in terms of distance from other levels in the buffer, $L$ is given by $$h({\cal D}_i, \left<{\cal D}_1, {\cal D}_2,\cdots\right>)$$ where ${\cal D}_i = {\cal D}(l_{\theta_i} L \setminus l_{\theta_i})$. We then convert the rank into a probability using Equation (\ref{eq:div_rank}),
\begin{eqnarray} 
P_{i} & = & \frac{1}{h({\cal D}_i, \mathbf{{\cal D}})^\beta} \label{eq:div_rank} 
\end{eqnarray}
where $h(\cdot, \cdot)$ is the rank transformation that can find the rank of ${\cal D}_i$ in a set of values $\cal D$, and $\beta$ is a tunable parameter.

\subsubsection{Combining Diversity with Regret/GAE}
Instead of solely considering the diversity aspect of the level buffer, we can also take learning potential into account by using regrets in Equation (\ref{eq:gae}) so that we will have a level buffer that is not only filled up with diverse training levels but also challenging levels that continuously push the student. We could assign different weights to diversity and regret by letting the replay probability $P_{replay}=\rho \cdot P_{D} + (1-\rho) \cdot P_{R}$, where $P_{D}$ and $P_{R}$ are the prioritization of diversity and regret respectively, and $\rho$ is the tuning parameter.

An overview of our proposed algorithm is shown in Figure \ref{fig:algo_pipeline}. When the level generator creates a new level $l_{\theta_i}$, we collect trajectories $\tau_i$ on $l_{\theta_i}$, and compute its distance ${\cal D}_i$ to the trajectory buffer by Equation (\ref{eq:div_level2buffer}). If the replay probability of $l_{\theta_i}$ is greater than that of any levels in the level buffer, we insert $l_{\theta_i}$ into the level buffer and remove the level with minimum replay probability from the level buffer.

The complete procedure of DIPLR is presented in Algorithm \ref{alg:algorithm1}. To accelerate the calculation process required by Wasserstein distance, we adopt the popular empirical Wasserstein distance solver $D(\cdot, \cdot)$ from \cite{flamary2021pot}. For simplicity, we use $\tau$ to represent the state-action samples from trajectory $\tau$ in the pseudocode. To better reveal the relationship between {\em diversity} and {\em learning potential}, we conduct an ablation study where we only adopt the diversity metric to pick levels to fill up the level buffer, i.e., set $P_{replay}=1 \cdot P_D + 0 \cdot P_R$.
\begin{algorithm}[t]
    \caption{DIPLR}
    \label{alg:algorithm1}
    \textbf{Input}: Level buffer size $N$, level generator $\mathcal{G}$ \\
    \textbf{Initialize}: student policy $\pi_\eta$, level buffer $L$, traj buffer $\Gamma$
    
    \begin{algorithmic}[1]

    \STATE Generate $N$ initial levels by $\mathcal{G}$ to populate $L$
    \STATE Collect trajectories on each replay level in $L$ and fill up $\Gamma$
        
        \WHILE{not converged}
        \STATE Sample replay-decision, $\epsilon \sim U[0,1]$
        \IF {$\epsilon \geq 0.5$}
        \STATE Generate a new level $l_{\theta_i}$ by $\mathcal{G}$
        \STATE Collect trajectories $\tau_i$ on $l_{\theta_i}$, with stop-gradient $\eta_{\perp}$

        
        \STATE Compute the regret, staleness and distance for $l_{\theta_i}$
        \ELSE
        \STATE Sample a replay level $l_{\theta_j}\in L$ according to $P_{replay}$
        \STATE Collect trajectories $\tau_j$ on $l_{\theta_j}$ and update $\pi_\eta$ with rewards $R(\tau_j)$
        \STATE Compute the regret, staleness and distance for $l_j$
        \ENDIF

        \STATE Flush $\Gamma$ and collect trajectories on all replay levels to fill up $\Gamma$
        \STATE Update regret, staleness, and distance for $l_{\theta_i}$ or $l_{\theta_j}$
        \STATE Update $L$ with new level $l_{\theta_i}$ if its replay probability is greater than any levels in $L$
    
        \STATE Update replay probability $P_{replay}$ 

        \ENDWHILE
    \end{algorithmic}
\end{algorithm}

\section{Experiments}
\label{experiments}
In this section, we compare DIPLR to the set of leading benchmarks for solving UED: Domain Randomization, Minimax, PAIRED, PLR$^\perp$. We conduct experiments and empirically demonstrate the effectiveness and generality of DIPLR on three popular yet highly distinct UPOMDP domains, Minigrid, Bipedal-Walker and Car-Racing. Minigrid is a partially-observable navigation problem under discrete control with sparse rewards, while Bipedal-Walker and Car-Racing are partially-observable walking/driving under continuous control with dense rewards (we provide more details in Appendix). In each domain, we train the teacher and student agents with Proximal Policy Optimization (PPO, \cite{schulman2017proximal}) and we will present the zero-shot out-of-distribution (OOD) test performance of all algorithms in each domain. To make the comparison more reliable and straightforward, we adopt the recently introduced standardized DRL evaluation metrics \cite{agarwal2021deep}, with which we show the aggregate inter-quartile mean (IQM) and optimality gap plots. Specifically, IQM discards the bottom and top 25\% of the runs and measures the performance in the middle 50\% of combined runs, and it is thus robust to outlier scores and is a better indicator of overall performance; Optimality gap captures the amount by which the algorithm fails to meet the "best performance", i.e., it assumes that a score (e.g., =1.0) is a desirable target beyond which improvements are not very important. We present the plots after normalizing the performance with a min-max range of solved-rate/returns for better visualization.  

We also provide a detailed ablation analysis to demonstrate the utility of diversity alone, by providing results with and without regret (or GAE). The version with diversity alone is referred to as DIPLR$^-$ and the one which uses both diversity and regret is referred to as DIPLR.

\begin{figure}[t]
  \centering
  \subfigure{\includegraphics[width=0.8\linewidth]{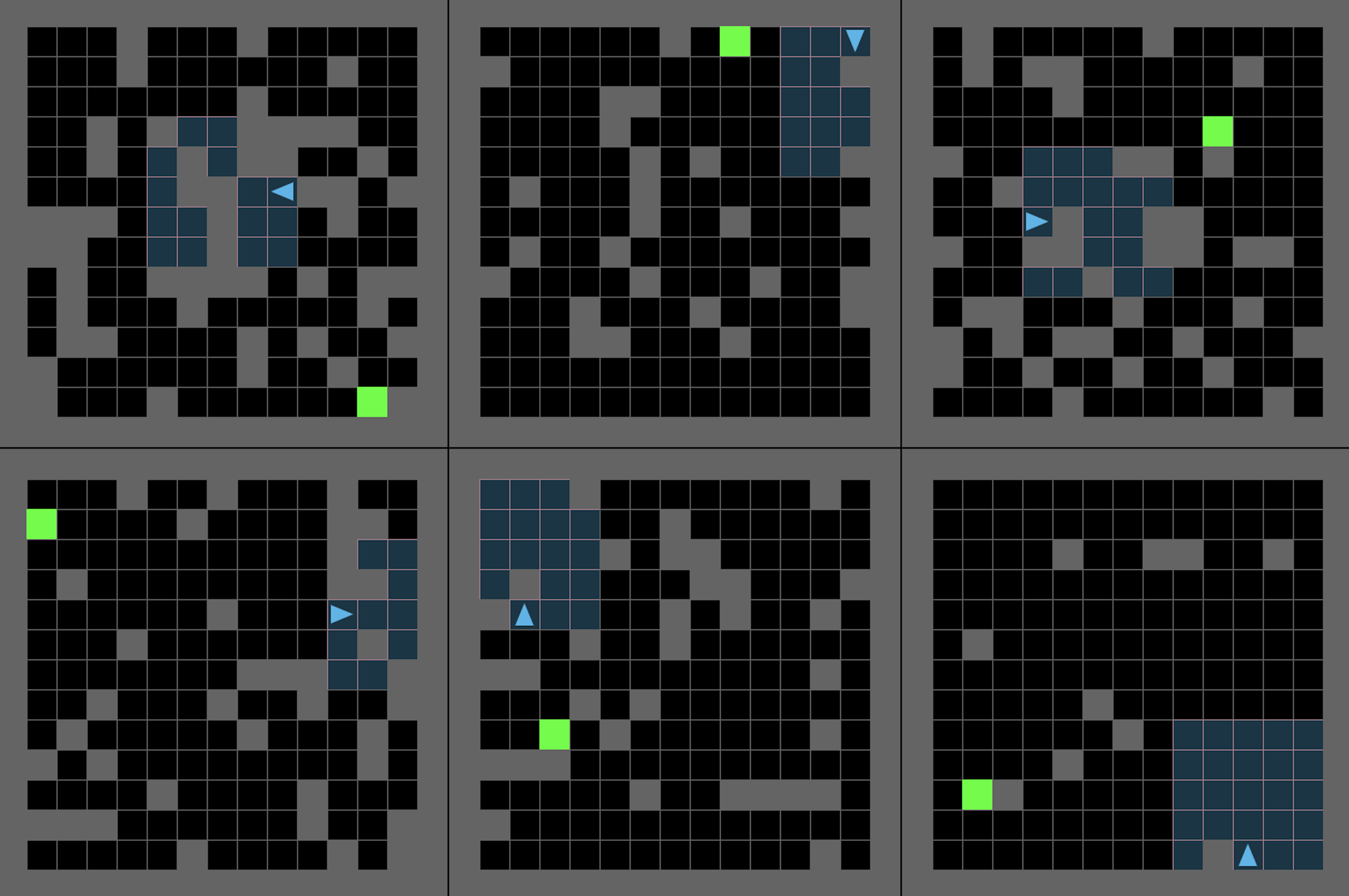}}
  \caption{Examples of training environments in Minigrid Domain}
  \label{fig:minigrid_domain}
\end{figure}
\begin{figure}[t]
  \centering
  \subfigure{\includegraphics[width=1.0\linewidth]{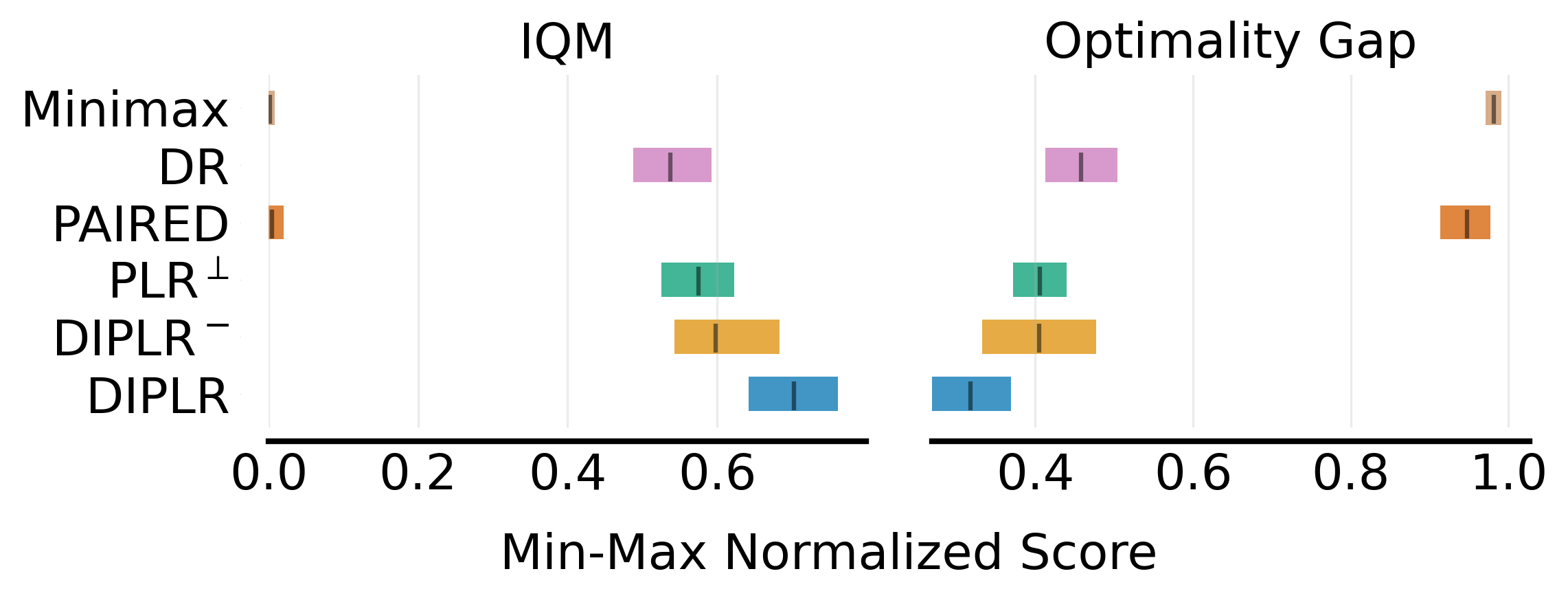}}
  \caption{Aggregate zero-shot OOD test performance in Minigrid Domain across 10 independent runs. Higher IQM scores and lower optimality gap are better.}
  \label{fig:results_mg_iqm}
\end{figure}
\begin{figure*}[t]
  \centering
  \subfigure{\includegraphics[width=1.0\linewidth]{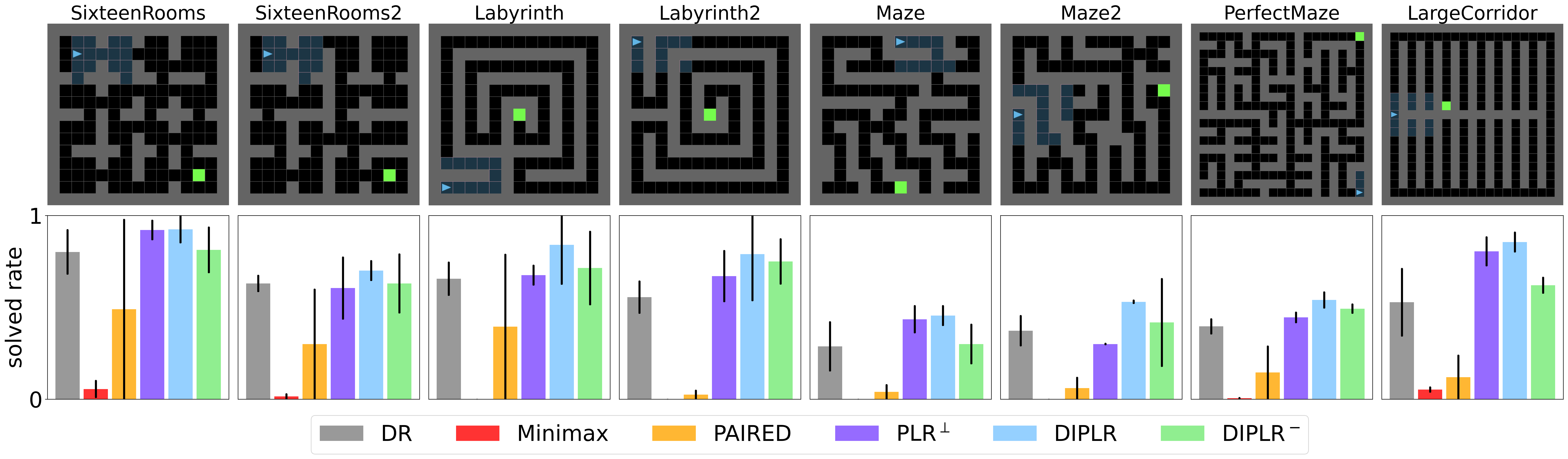}}
  \caption{Zero-shot transfer performance in eight human-designed test environments. The plots are based on the median and interquartile range of solved rates across 10 independent experiments.}
  \label{fig:results_mg}
\end{figure*}

\subsection{Minigrid Domain}
Introduced by \cite{dennis2020emergent}, Minigrid is a good benchmark due to its easy interpretability and customizability. In our experimental settings, the teacher has a maximum budget of 60 steps to place small block cells and another two steps to place the start position and goal position at the end. The student agent is required to navigate to the goal cell within 256 steps and it receives rewards=(\textit{num-steps}/256) upon success, otherwise, it receives zero rewards. Figure \ref{fig:minigrid_domain} displays a few training examples from the domain.

We train all the student agents for 30k PPO updates ($\sim$250M steps) and evaluate their transfer capabilities on eight zero-shot OOD environments (see the first row in Figure \ref{fig:minigrid_domain}) used in the literature that are crafted by humans. We summarize the empirical results of our proposed algorithm and all other benchmarks in Figure \ref{fig:results_mg_iqm} and Figure \ref{fig:results_mg}. Here are the key observations from both the figures:

\begin{itemize}
    \item Figure \ref{fig:results_mg_iqm}: DIPLR surpasses all the benchmark approaches, and surpasses the leading best approach, PLR$^{\bot}$ by more than 20\% with regards to IQM and Optimality Gap.
    \item Figure \ref{fig:results_mg_iqm}: DIPLR$^-$ performs on par or better than PLR$^{\bot}$.
    \item Figure \ref{fig:results_mg_iqm}: PAIRED performs poorly, as the long horizon (60 steps) and sparse reward task is challenging for a guided level generator. While Domain Randomization performs decently, minimax performs worse than PAIRED. 
    \item Figure \ref{fig:results_mg}: With respect to zero-shot transfer results, DIPLR is obviously better than PLR$^\bot$ in every one of the testing scenarios. This indicates the utility of diversity. 
\end{itemize}

\begin{figure}[t]
  \centering
  \subfigure{\includegraphics[width=1\linewidth]{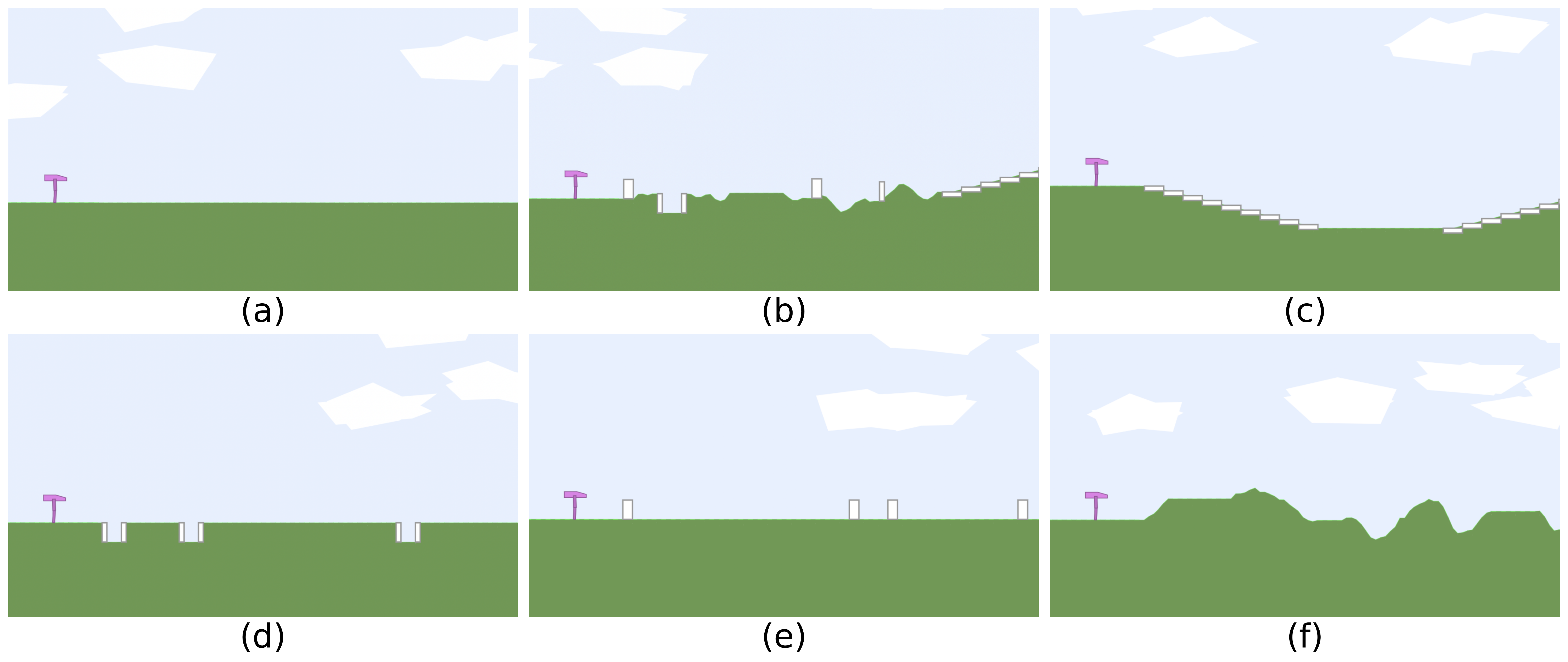}}
  \caption{Six examples of test environments in Bipedal-Walker domain. (a) BipedalWalker, (b) Hardcore, (c) Stair, (d) PitGap, (e) Stump, (f) Roughness. Note that these may not be the exact test environment terrains but only extreme cases for demonstration.}
  \label{fig:bw_domain}
\end{figure}
\begin{figure}[t]
  \centering
  \subfigure{\includegraphics[width=1.0\linewidth]{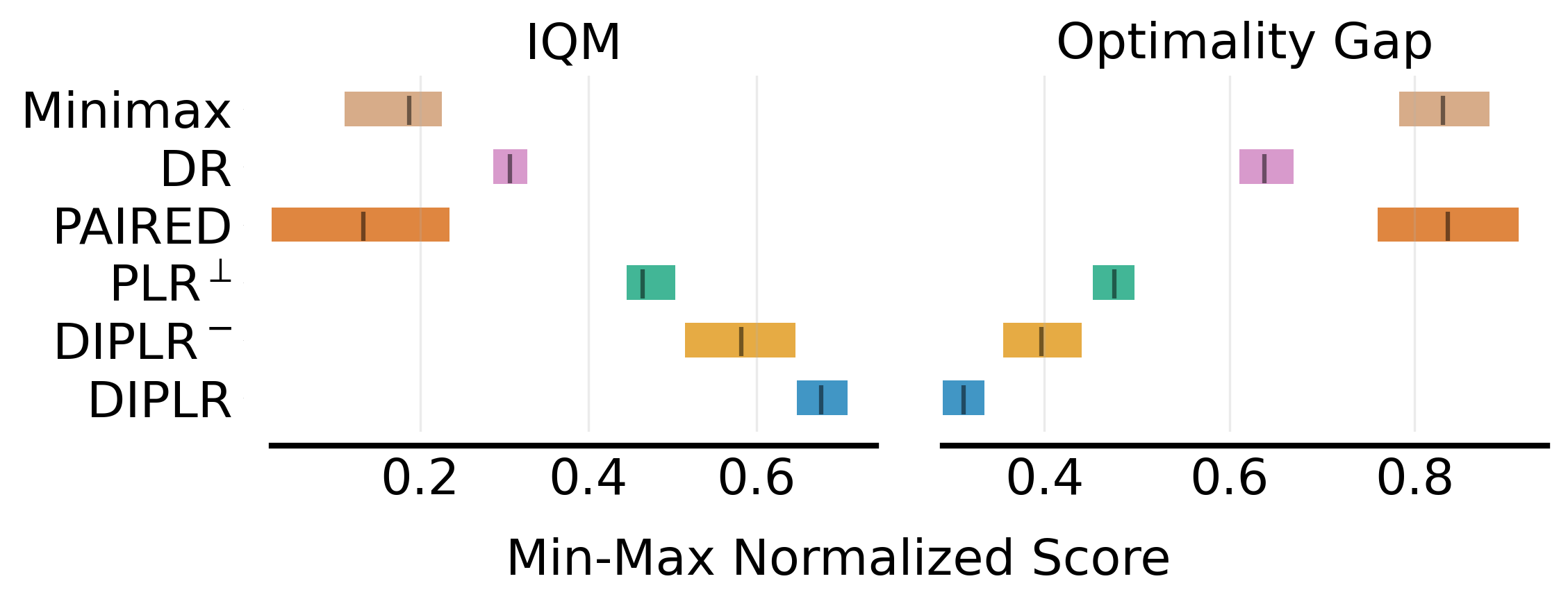}}
  \caption{Aggregate test performance over 10 runs in Bipedal-Walker domain. Higher IQM scores and lower optimality gap are better}
  \label{fig:results_bw_iqm}
\end{figure}
\begin{figure*}[ht]
  \centering
  \subfigure{\includegraphics[width=7.0in, height=1.5in]{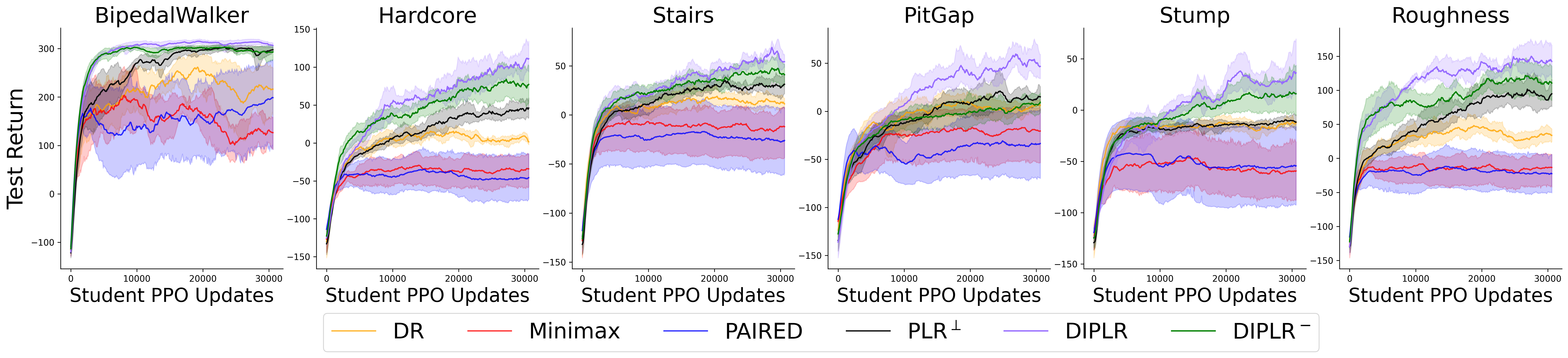}}
  \caption{Performance on test environments during the training period over five independent experiments (mean and standard error).}
  \label{fig:results_bw}
\end{figure*}

\subsection{Bipedal-Walker Domain}
We also conduct a comprehensive comparison on the Bipedal-Walker domain, which is introduced in \cite{wang2019paired,parker2022evolving} and is popular in the community as it is a well-parameterized environment with eight variables (including ground roughness, the number of stair steps, min/max range of pit gap width, min/max range of stump height, and min/max range of stair height) conditioning the terrains and is thus straightforward to interpret. The teacher in Bipedal-Walker can specify the values of the eight environment parameters, but there will still be stochasticity in the generation of a particular level. As for the student, i.e., walker agent, it should determine the torques applied on its joints and is constrained by partial observability where it only knows its horizontal speed, vertical speed, angular speed, positions of joints, joints' angular speed, etc. 


Following the experiment settings in prior UED works, we train all the algorithms for 30k PPO updates ($\sim$1B steps), and then evaluate their generalization ability on six distinct test instances in Bipedal-Walker domain, i.e., BidedalWalker, Hardcore, Stair, PitGap, Stump, and Roughness, as illustrated in Figure~\ref{fig:bw_domain}. Among these, $\left\{BipedalWalker\right\}$  is the basic level that only evaluates whether the agent can walk on flat ground, $\left\{Stair, PitGap, Stump, Roughness \right\}$ challenges the walker for a particular kind of obstacles and $\left\{Hardcore\right\}$ contains a combination of all types of challenging obstacles. 

To understand the evolution of transfer performance, we evaluate the student every 100 student policy updates and plot the curves in Figure \ref{fig:results_bw}. As is shown, our proposed method DIPLR outperforms all other benchmarks in each test environment. The ablation model DIPLR$^-$ also surpasses our primary benchmark PLR$^\perp$ in five test environments, indicating that diversity metric alone contributes to ultimate generalization performance more than regret in continuous domains. PAIRED suffers from variance quite a bit and cannot constantly create informative levels for the student agents in such a complex domain. This is because of the nature of the multi-agent framework adopted in PAIRED, where convergence highly depends on the ability of the expert student.

After training for 30k PPO updates, we collect trained models and conduct a more rigorous evaluation based on 10 test episodes in each test environment, and present the aggregate results after min-max normalization (with range=[0, 300] on Bipedal-Walker and [0, 150] on the other five test environments) in Figure \ref{fig:results_bw_iqm}. Notably, our method DIPLR dominates all the benchmarks in both IQM and optimality gap. Furthermore, DIPLR$^-$ performs better than PLR$^\bot$. 

\begin{figure}[t]
  \centering
  \subfigure{\includegraphics[width=1.0\linewidth]{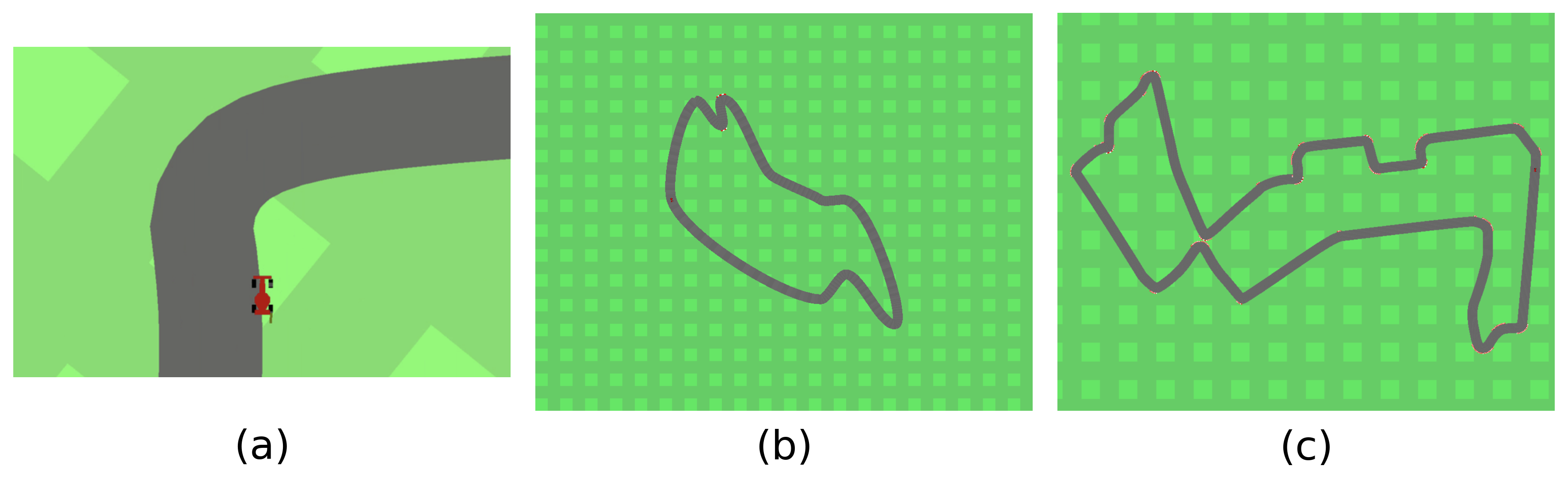}}
  \caption{Examples in Car-Racing domain. (a) A zoom-in snapshot of the car-racing track; (b) a track generated by domain randomization; (c) one of the test environments, F1-Singapore track }
  \label{fig:cr_domain}
\end{figure}
\begin{figure}[t]
  \centering
  \subfigure{\includegraphics[width=3.4in, height=1.25in]{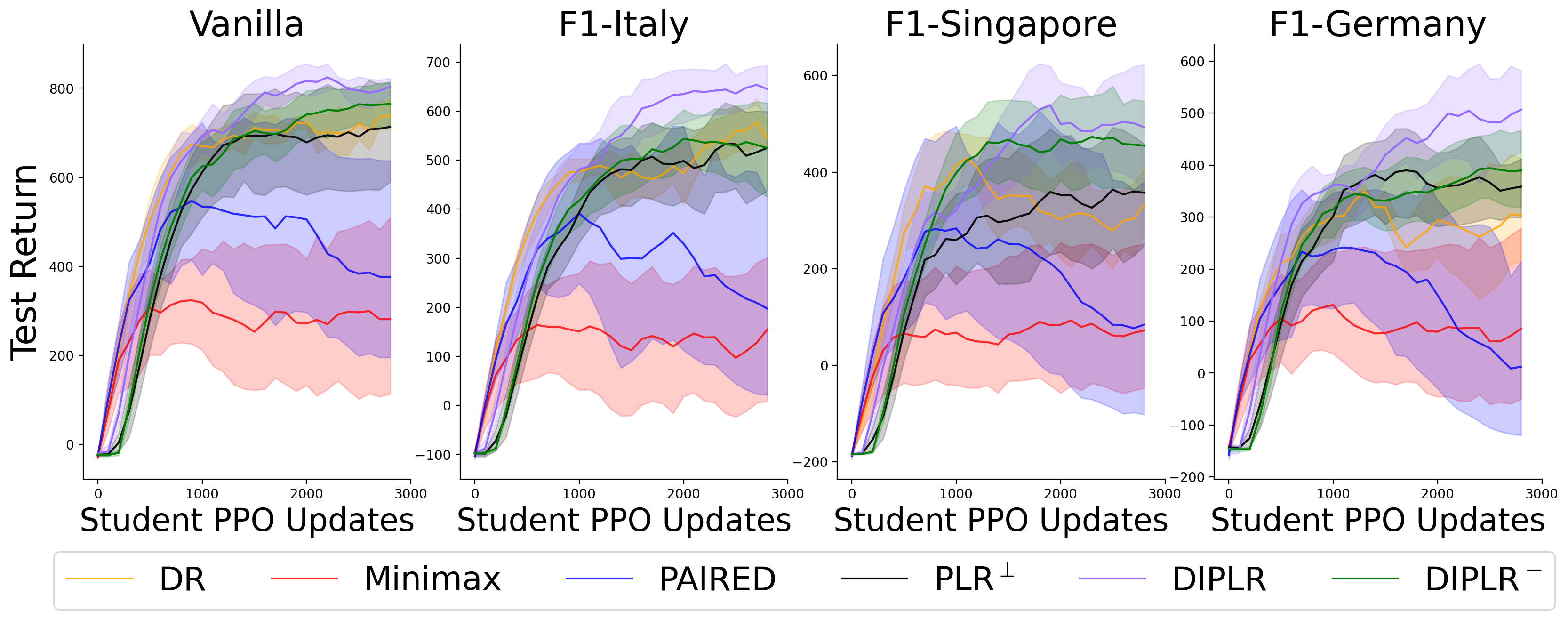}}
  \caption{Zero-Shot Transfer Performance in Car-Racing domain}
  \label{fig:results_cr}
\end{figure}
\begin{figure}[t]
  \centering
  \subfigure{\includegraphics[width=3.3in, height=1.1in]{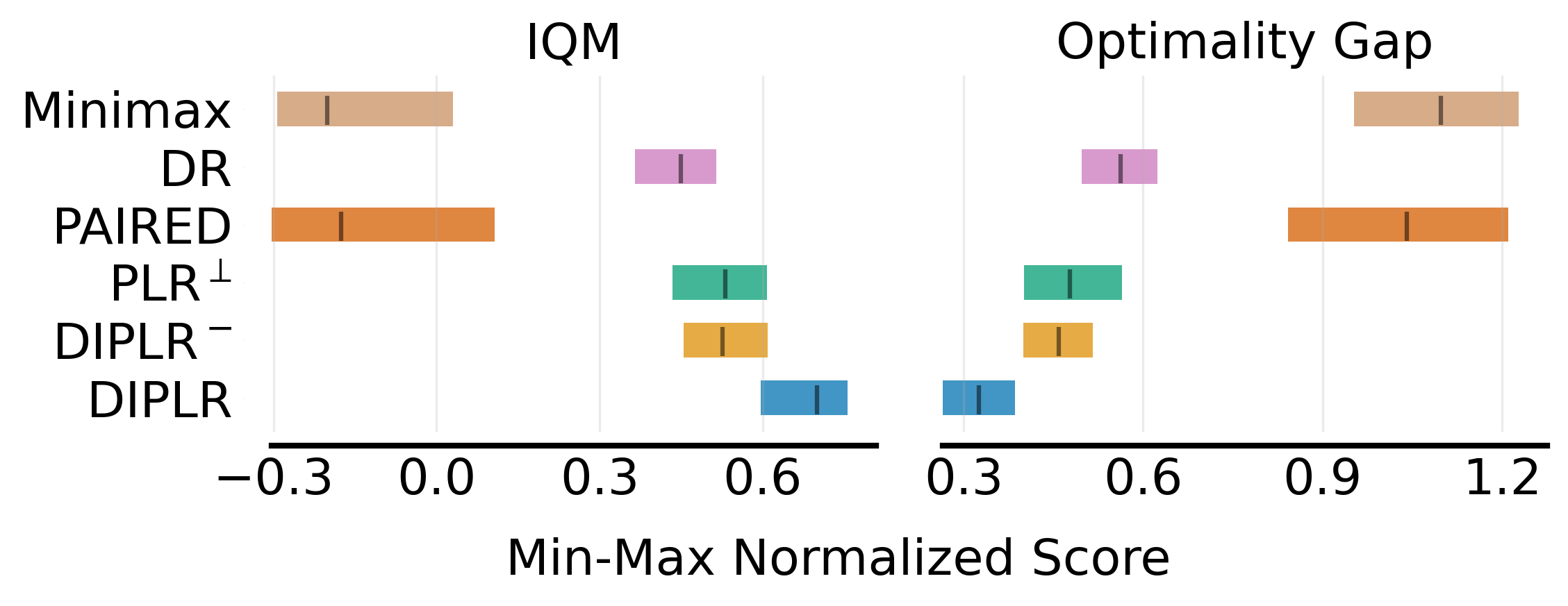}}
  \caption{Aggregate test performance over 10 independent runs in Car-Racing domain.}
  \label{fig:results_cr_iqm}
\end{figure}

\subsection{Car-Racing Domain}
To validate that our method is scalable and versatile, we further implement DIPLR on the Car-Racing environment, which is introduced by \cite{jiang2021replay} and has been used in existing UED papers. In this domain, the teacher needs to create challenging racing tracks by using B{\'e}zier curves (via 12 control points) and the student drives on the track under continuous control with dense reward. A zoom-in illustration of the track is shown in Figure \ref{fig:cr_domain} (a). After training the student for around 3k PPO updates ($\sim$ 5.5M steps), we evaluate the student agent in four test scenarios, among which three are challenging F1-tracks existing in the real world and the other one is a basic vanilla track. Note that these tracks are absolutely OOD because they cannot be defined with just 12 control points, for example, the F1-Singapore track in Figure \ref{fig:cr_domain} (c) has more than 20 control points. 

Figure \ref{fig:results_cr} presents the complete training process of each algorithm on four test environments, with an evaluation interval of 100 PPO updates. As a relatively simpler domain, our method DIPLR and DIPLR$^-$ quickly converge to the optimum while DR and PLR$^\perp$ achieve a near-optimal generalization on the Vanilla and F1-Italy track. 

The aggregate performance after min-max normalization (with range=[200,800]) of all methods is summarized in Figure \ref{fig:results_cr_iqm}. Despite both PLR$^\perp$ and DR agents can learn well on this comparatively simple domain, DIPLR outperforms them on both IQM and optimality gap.

\section{Conclusion}
We provided a novel method DIPLR for unsupervised environment design (UED), where diversity in training environments is employed to improve generalization ability of a student agent. By computing the distance between levels via Wasserstein distance over occupancy distributions, we constantly expose the agent to challenging yet diverse scenarios and thus improve its generalization in zero-shot out-of-distribution test environments. In our experiments, we validated that DIPLR is capable of training robust and generalizing agents and significantly outperforms the best-performing baselines in three distinct UED domains. Moreover, we explored the relationship between {\em diversity} and {\em learning potential}, and we discover that diversity alone benefits the algorithm more than learning potential, and they complement each other to achieve the best performance when combined. 



\section*{Acknowledgments}
This research/project is supported by the National Research Foundation Singapore and DSO National Laboratories under the AI Singapore Programme (AISG Award No: AISG2-RP-2020-017).

\bibliographystyle{named}
\bibliography{ijcai23}

\end{document}